\def\paperlanguage{}
\pdfoutput=1 % for arxiv

\documentclass[letterpaper, 10 pt, conference]{ieeeconf}  % Comment this line out if you need a4paper

\usepackage{bm}
\usepackage{cite}
\include{preamble}

\newcommand{\ctext}[1]{\raise0.2ex\hbox{\textcircled{\scriptsize{#1}}}}

\IEEEoverridecommandlockouts                              % This command is only needed if
\title{\LARGE \textbf
  {
    \switchlanguage%
    {%
      Continuous Object State Recognition for Cooking Robots Using Pre-Trained Vision-Language Models and Black-box Optimization
    }%
    {%
      事前学習済み視覚-言語モデルとブラックボックス最適化を応用した\\調理ロボットのための時系列物体状態認識
    }%
  }
}

\author{Kento Kawaharazuka$^{1}$, Naoaki Kanazawa$^{1}$, Yoshiki Obinata$^{1}$, Kei Okada$^{1}$, and Masayuki Inaba$^{1}$% <-this % stops a space
  \thanks{$^{1}$ The authors are with the Department of Mechano-Informatics, Graduate School of Information Science and Technology, The University of Tokyo, 7-3-1 Hongo, Bunkyo-ku, Tokyo, 113-8656, Japan.
    {\texttt\small [kawaharazuka, kanazawa, obinata, k-okada, inaba]@jsk.t.u-tokyo.ac.jp}
  }
}
\begin{document}

\maketitle
\thispagestyle{empty}
\pagestyle{empty}

%%%%%%%%%%%%%%%%%%%%%%%%%%%%%%%%%%%%%%%%%%%%%%%%%%%%%%%%%%%%%%%%%%%%%%%%%%%%%%%%
\begin{abstract}
  \switchlanguage%
  {%
    The state recognition of the environment and objects by robots is generally based on the judgement of the current state as a classification problem.
    On the other hand, state changes of food in cooking happen continuously and need to be captured not only at a certain time point but also continuously over time.
    In addition, the state changes of food are complex and cannot be easily described by manual programming.
    Therefore, we propose a method to recognize the continuous state changes of food for cooking robots through the spoken language using pre-trained large-scale vision-language models.
    By using models that can compute the similarity between images and texts continuously over time, we can capture the state changes of food while cooking.
    We also show that by adjusting the weighting of each text prompt based on fitting the similarity changes to a sigmoid function and then performing black-box optimization, more accurate and robust continuous state recognition can be achieved.
    We demonstrate the effectiveness and limitations of this method by performing the recognition of water boiling, butter melting, egg cooking, and onion stir-frying.
  }%
  {%
    ロボットによる環境や物体の状態認識は, 現在の状態を分類問題として判断することが一般的である.
    一方で, 料理に代表されるような食材の状態変化は連続的であり, ある時刻だけでなく, それを連続的な時間方向で捉える必要がある.
    また, その状態変化は複雑であり, 容易にプログラミングによって記述できるようなものではない.
    そこで本研究では, 料理ロボットに向けた食材の連続状態認識を, 事前学習済みの大規模視覚-言語モデルを用いて, 言語を通して行う手法を提案する.
    画像と言語テキストの類似度を計算可能なモデルを時間方向に連続的に用いることで, 食材の状態変化を捉える.
    また, 類似度変化のシグモイド関数へのフィッティングとブラックボックス最適化に基づき, 各言語テキストに対する重みを調整することで, より正確でロバストな連続状態認識が可能になることを示す.
    水の沸騰認識, バターの解け具合認識, 卵の固まり具合認識, 玉ねぎの炒め具合認識を行い, その有効性と限界を示す.
  }%
\end{abstract}

\section{INTRODUCTION}\label{sec:introduction}
\switchlanguage%
{%
  State recognition of the environment and objects by robots is essential for various tasks such as daily life support, security, and disaster response.
  This includes recognition of the open/closed state of doors, the on/off state of lights, and the relationships among objects, all of which determine the state of objects at a certain time \cite{chin1986recognition, quintana2018door, kawaharazuka2023ofaga}.
  On the other hand, changes in the state of food, as represented by cooking, happen continuously and need to be captured not only at a certain time point but also continuously over time.
  In addition, the state changes are complex and cannot be easily described by manual programming.
  Even if each state recognition is trained by a neural network, it is difficult to cover the wide variety of state changes that occur in food, and it is necessary to prepare datasets, models, and programs for each state recognition, which also creates problems in managing source codes and computational resources.

  Various cooking robots have been developed to handle changes in the state of food.
  \cite{beetz2011pancake} has developed a system where two robots cook pancakes from a recipe.
  \cite{junge2020omelette} has developed a system to optimize the quality of omelette cooking by batch bayesian optimization.
  However, these two systems do not capture the state changes of food directly, and the cooking is based on the time of applying heat.
  Although this has a certain effect, it is important to capture changes in the state of food directly, taking into account the individual differences of the food, the differences in heat power, and cooking with unknown recipes.
  On the other hand, there have been some studies that capture changes in the state of food by using images \cite{paul2018cooking, jelodar2018cooking, sakib2021cooking, takata2022cooking}.
  However, these all deal with classification problems based on convolutional neural networks, which determine whether the vegetable is sliced, shredded, chopped, etc., and cannot capture continuous changes in the state of food.
  At the same time, since the recognition is based on a predefined classification, it is difficult to respond to changes in states that are not included in the classification.
  It is also difficult to understand the degree of change.
}%
{%
  ロボットによる環境や物体の状態認識は, 日常生活支援や警備, 災害対応等の様々なタスクにおいて欠かせない.
  これには, ドアの開閉状態や電気のオンオフ, 物体同士の関係性などの認識が含まれるが, そのどれもが現在の一状態を判断するものである\cite{chin1986recognition, quintana2018door, kawaharazuka2023ofaga}.
  一方で, 料理に代表されるような食材の状態変化は連続的であり, ある一時刻だけでなく, それを連続的な時間方向で捉える必要がある.
  また, 加熱調理に代表される状態変化は複雑であり, 容易にプログラミングによって記述できるようなものではない.
  一つ一つの状態認識をニューラルネットワークによって学習しても, 食材に起こる多様な状態変化を網羅することは困難であり, 各状態認識に対してデータセットやモデル, プログラムを用意する必要があるため, ソースコードや計算リソースの管理にも問題が生じる.

  これまで, 食材の状態変化を扱う様々な料理ロボットが開発されてきた.
  \cite{beetz2011pancake}は二台のロボットでレシピ記述からパンケーキを調理するシステムを開発した.
  \cite{junge2020omelette}は, オムレツ調理のクオリティをbatch bayesian optimizationにより最適化するシステムを開発した.
  しかし, 上記2つは状態変化を直接捉えておらず, 熱を加えた時間に基づき調理を行っている.
  もちろんこれは一定の効果があるが, 食材の個体差や火力の違い, 未知のレシピに対する調理を考慮に入れると, 食材の状態変化を直接捉えることは重要である.
  これに対して, 食材の状態変化を画像により捉えた研究も行われてきている\cite{paul2018cooking, jelodar2018cooking, sakib2021cooking, takata2022cooking}.
  しかし, これらは全てConvolutional Neural Networkによる分類問題を扱っており, その野菜がスライスされたものか, 千切りされたものか, みじん切りにされたものか等を判定するものであり, 食材の連続的な状態変化を捉えることはできない.
  同時に, 予め定めた分類に基づく認識のため, 分類にない状態変化に対応することは困難であり, 変化の程度を知ることも難しい.
}%

\begin{figure}[t]
  \centering
  \includegraphics[width=0.95\columnwidth]{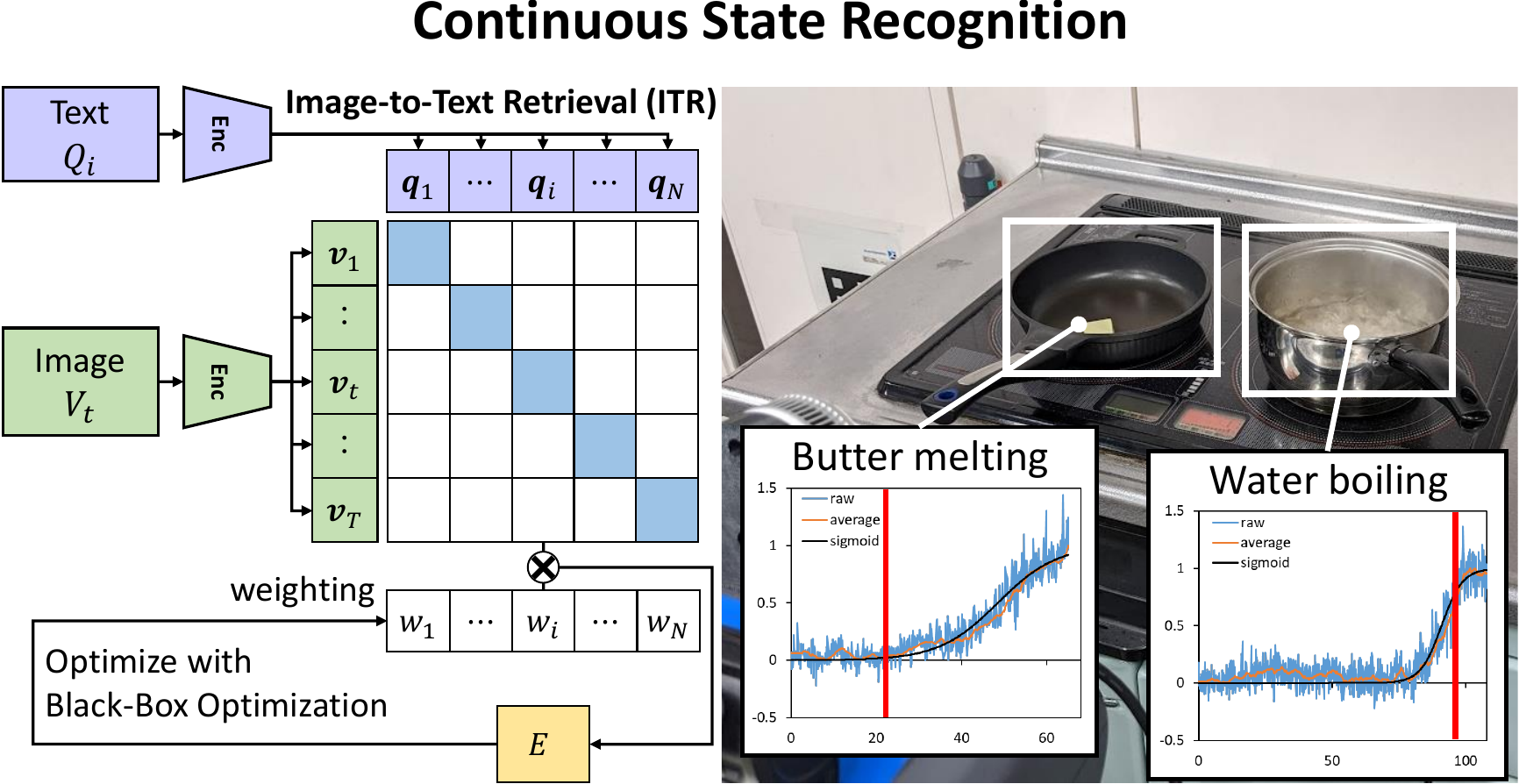}
  \vspace{-1.0ex}
  \caption{The concept of this study. We propose a continuous object state recognition method for cooking robots by using pre-trained large-scale vision-language models and black-box optimization.}
  \label{figure:concept}
  \vspace{-3.0ex}
\end{figure}

\switchlanguage%
{%
  Therefore, we propose a method to continuously recognize changes in the state of food for cooking robots through the spoken language using pre-trained large-scale vision-language models (VLMs) \cite{li2022largemodels, kawaharazuka2023ptvlm} as shown in \figref{figure:concept}.
  % In this study, among the tasks for which VLMs are possible, we focus on a task that computes the similarity between images and texts.
  In this study, we utilize VLMs that have learned the semantic correspondences between images and the spoken language through a large dataset \cite{radford2021clip, girdhar2023imagebind}.
  By using the spoken language, the proposed method can appropriately capture diverse and ambiguous state changes in the cooking process.
  Due to the semantic training of correspondences, VLMs are also robust to changes in images.
  Moreover, by using pre-trained VLMs, the method does not require any manual programming or training of neural networks.
  Using only a single VLM makes it easy to manage the source codes and computational resources for each state to be recognized.
  We prepare a set of various texts about the state of the food to be recognized, and capture the state changes as continuous changes in the similarity between the texts and the current image.
  We also show that by adjusting the weighting of each text prompt based on fitting the similarity changes to a sigmoid function and then performing black-box optimization, more accurate and robust continuous state recognition can be achieved.
  The sigmoid function is capable of representing the basic patterns of continuous state changes through its parameter variations and is well-suited for continuous state recognition.
  Our method corresponds to obtaining a value that changes more significantly in synchronization with the state change.
  It is important to note that no annotation of images is required since only the degree of change is captured.
  We demonstrate the effectiveness and limitations of our method through experiments on water boiling, butter melting, egg cooking, and onion stir-frying.
  Our contributions are summarized as follows:
  \begin{itemize}
    \item \textbf{New Continuous State Recognition Task}: Proposing a continuous recognition task for the state changes of food for cooking robots.
    \item \textbf{Capturing Cooking State Changes}: Capturing diverse and ambiguous state changes during the cooking process through spoken language analysis.
    \item \textbf{Simplified Implementation}: Eliminating the need for manual programming or neural network training, and ensuring easy code management and efficient resource use with a single pre-trained vision-language model.
    \item \textbf{Improved Performance}: Achieving accurate continuous state recognition by adjusting text prompt weights using a sigmoid function and black-box optimization.
    % \item \textbf{Method Evaluation}: Demonstrating the effectiveness of our approach through experiments on water boiling, butter melting, egg cooking, and onion stir-frying.
  \end{itemize}
}%
{%
  そこで本研究では, 料理ロボットに向けた食材の連続状態認識を, 事前学習済みの大規模視覚-言語モデル\cite{li2022largemodels, kawaharazuka2023ptvlm}を用いて, 言語を通して行う手法を提案する(\figref{figure:concept}).
  本研究では, 大規模視覚-言語モデルが可能なタスクの中でも, 画像と言語テキストの類似度を計算するタスクに着目する.
  大規模なデータをもとに画像と自然言語の間の意味的な相関を学習した視覚-言語モデルを利用する\cite{radford2021clip, girdhar2023imagebind}.
  自然言語を用いることで, 料理中の多様で曖昧な状態変化を適切に捉えることができる.
  また, 事前学習済みの大規模言語モデルを用いるため, 一切の手動プログラミングやニューラルネットワークの学習を必要としない.
  たった一つの視覚言語モデルのみ用いるため, 認識したい状態ごとのもソースコードや計算リソースの管理も容易である.
  認識したい状態に関する多様なテキスト集合を用意し, これと現在画像の間の類似度の連続的な変化によって, 食材の状態変化を捉える.
  また, 類似度変化のシグモイド関数へのフィッティングとブラックボックス最適化に基づき, 各言語テキストに対する重みづけを調整することで, より正確でロバストな連続状態認識が可能になることを示す.
  シグモイド関数は, そのパラメータ変化によって, 基本的な連続状態変化のパターンを表現可能であり, 連続状態認識に利用するのに適している.
  これは, ある状態変化と同期してより大きく変化する状態量を得ることに相当し, 変化度合いのみを捉えるため画像のアノテーションは一切必要ない点も重要である.
  水の沸騰認識, バターの解け具合認識, 卵の固まり具合認識, 玉ねぎの炒め具合認識を行い, 本手法の有効性と限界を示す.
}%

\section{Robotic Continuous State Recognition Using Pre-Trained Vision-Language Models and Black-box Optimization} \label{sec:proposed}
\subsection{Pre-Trained Vision-Language Models for Robotic Continuous State Recognition} \label{subsec:models}
\switchlanguage%
{%
  There are various types of pre-trained VLMs.
  Among them, it is necessary to obtain the results as continuous values rather than discrete ones for continuous state recognition.
  \cite{li2022largemodels} has classified the tasks that VLMs can handle into four categories: Generation Task, Understanding Task, Retrieval Task, and Grounding Task.
  Generation Task includes Image Captioning (IC) and Text-to-Image Generation (TIG).
  Understanding Task includes Visual Question Answering (VQA), Visual Dialog (VD), Visual Reasoning (VR), and Visual Entailment (VE).
  Retrieval Task includes Image-to-Text Retrieval (ITR) and Text-to-Image Retrieval (TIR), which retrieve the correspondence between images and texts from alternatives using similarity.
  Grounding Task includes Visual Grounding (VG), which extracts the corresponding regions in the image from the text.
  Among these tasks, only ITR and VG output continuous numerical values, while the others output sentences or images.
  Between the two, only ITR, which can compute the similarity between the current image and the texts describing the change in the state of food, is consistent with our purpose.

  In this study, we conduct experiments using CLIP \cite{radford2021clip} and ImageBind \cite{girdhar2023imagebind} as models that are capable of ITR.
  CLIP is a model that can calculate the cosine similarity between images and texts by vectorizing them into latent space.
  ImageBind is a model that can compute similarity not only for images and texts, but also for many other modalities including audio, depth images, heatmaps, and inertial sensors.
  Note that it is necessary to extract the target region for state recognition, which can be done by using VG provided in \cite{wang2022ofa}.
}%
{%
  事前学習済みの大規模視覚-言語モデルには様々な種類がある.
  その中でも, 連続的に状態認識をするためには, 結果を離散的ではなく連続的な数値として得る必要がある.
  \cite{li2022largemodels}では, 大規模視覚-言語モデルが可能なタスクをGeneration Task, Understanding Task, Retrieval Task, Grounding Taskの4つに分類している.
  Generation TaskにはImage Captioning (IC)やText-to-Image Generation (TIG)が含まれる.
  Understanding Taskには, Visual Question Answering (VQA), Visual Dialog (VD), Visual Reasoning (VR), Visual Entailment (VE)が含まれる.
  Retrieval Taskには, 画像と言語の対応を類似度を用いて選択肢から検索するImage-to-Text Retrieval (ITR)とText-to-Image Retrieval (TIR)が含まれる.
  Grounding Taskには, 言語から画像中の対応する箇所の領域を抜き出すVisual Grounding (VG)が含まれる.
  これらの中で, ITRとVGのみが連続的な数値を出力するタスクであり, その他は文章または画像を出力するタスクである.
  その中でも, 食材の変化を見るためには, 現在の画像とその状態変化を表現する言語テキストの類似度が計算可能なITRのみが本研究の意図に沿う.

  本研究では, ITRが可能なモデルとして具体的に, CLIP \cite{radford2021clip}とImageBind \cite{girdhar2023imagebind}を用いて実験を行う.
  CLIPは画像とテキストを用意し, これらをそれぞれベクトル化, それらの間のコサイン類似度を計算可能なモデルである.
  ImageBindは, 画像やテキストのみでなく, 音声や深度画像, ヒートマップや慣性センサを含む多数のモーダルについて同様に類似度が計算可能なモデルである.
  なお, 状態認識の際に対象領域を切り出す必要があるが, これには\cite{wang2022ofa}に備わるVGを使うことで可能である.
}%

\begin{figure}[t]
  \centering
  \includegraphics[width=0.95\columnwidth]{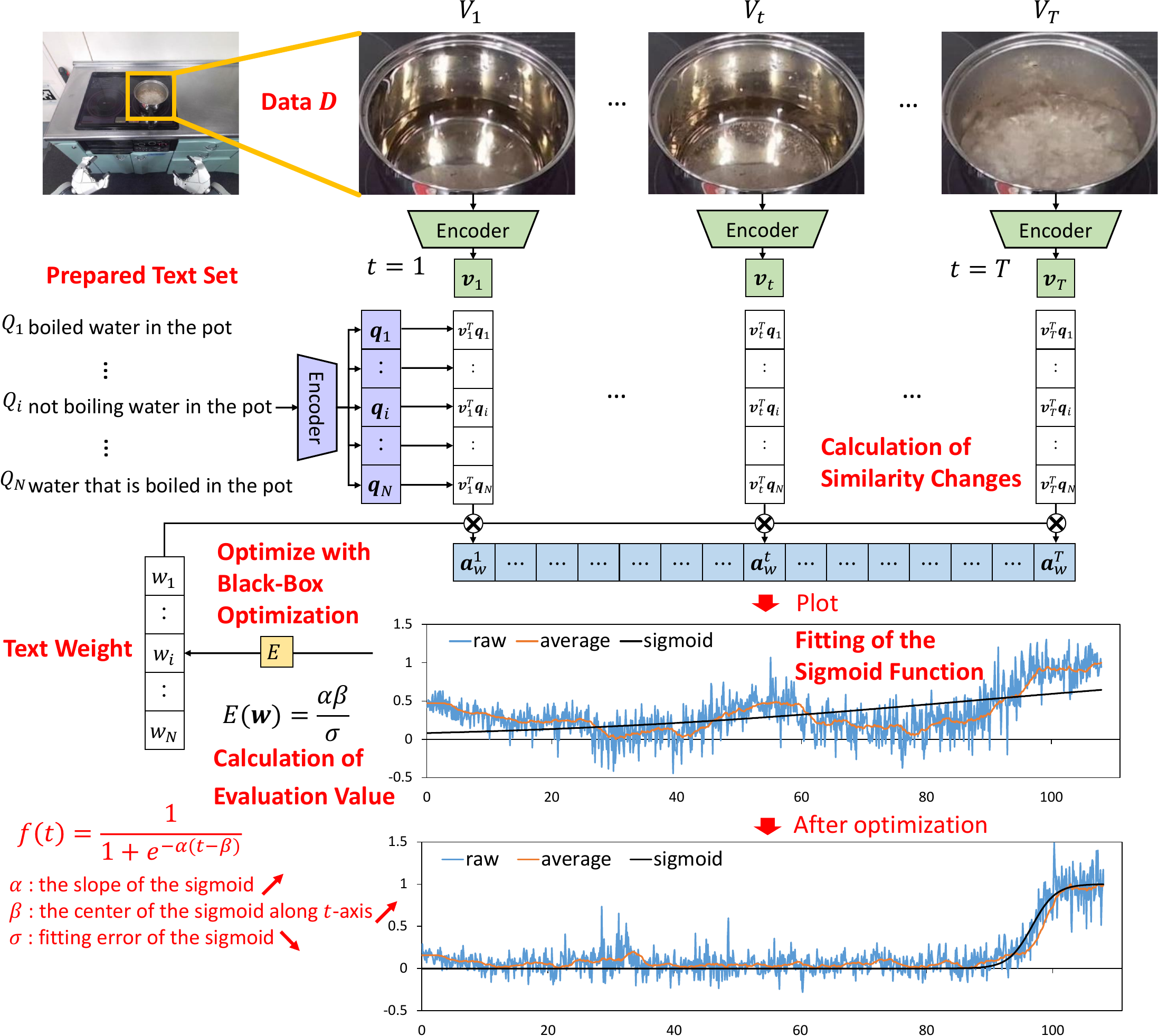}
  \vspace{-1.0ex}
  \caption{The overview of the proposed method: we obtain time series of images $D$, prepare a variety of text prompts, calculate the continuous similarity changes with pre-trained vision-language models and text weight, fit the similarity changes to a sigmoid function, compute the evaluation value, and iteratively optimize the text weight with black-box optimization.}
  \label{figure:proposed}
  \vspace{-3.0ex}
\end{figure}

\subsection{Robotic Continuous State Recognition Using Pre-Trained Vision-Language Models} \label{subsec:state-recognition}
\switchlanguage%
{%
  Continuous state recognition is performed using VLMs capable of ITR.
  The method is simple.
  First, we prepare a text prompt $Q$ for the state to be recognized.
  For example, ``boiled water'' to recognize water boiling and ``melted butter'' to recognize butter melting.
  Image $V$ is continuously acquired (at 10 Hz in this study), $V$ and $Q$ are vectorized into $\bm{v}$ and $\bm{q}$ using ITR, respectively, and the cosine similarity $\bm{v}^{T}\bm{q}$ is calculated.
  By plotting them continuously over time, we can quantify continuous state changes.
  In the case of water boiling recognition, the similarity of the current image to the text ``boiled water'' gradually increases.
  If we obtain the moving average of the similarity, we can recognize the beginning of the state change when the slope of the value with respect to time becomes large, or we can recognize the end of the state change when the slope becomes small.
  It is also possible to set a certain threshold and recognize that a state change has started or ended when the value exceeds the threshold.

  The text input to VLMs does not have to be a single text, but can be multiple synonyms and antonyms (in the case of antonyms, it is necessary to add a minus sign, $-\bm{v}^{T}\bm{q}$).
  It is also possible to add noise to the current image and take the average, or to use multiple models.
  In \cite{kanazawa2023cooking}, we have experimented with the case where a set of antonyms is prepared and one model is used.
}%
{%
  ITRが可能な大規模視覚-言語モデルを用いて連続状態認識を行う.
  方法は非常に単純である.
  まず, 認識した状態に関する言語プロンプト$Q$を用意する.
  例えば, 水の沸騰を認識したければ``boiled water'', バターの解け具合を認識したければ``melted butter''などである.
  画像$V$を連続的に得ていき(本研究では10 Hzで画像を取得する), $V$と$Q$をそれぞれITRが可能なモデルにより$\bm{v}$と$\bm{q}$にベクトル化, これらのコサイン類似度$\bm{v}^{T}\bm{q}$を計算する.
  これを時間方向に連続的にプロットしていくことで, 連続的な状態変化を得ることができる.
  水の沸騰認識であれば, 現在画像の``boiled water''に対する類似度が徐々に上昇していくことになる.
  例えばその移動平均を得たとして, 時間に対する状態変化の傾きが大きくなったら状態変化が始まったことを認識したり, 傾きが小さくなったら状態変化が終了したことを認識したりすることが可能である.
  また, ある設定した閾値を設けて, それを越えたら状態変化が始まった, または終わったと認識しても良い.

  本手法は様々な形で認識性能を変化させることができる.
  言語テキストはたった一つのテキストである必要はなく, 複数の同義語を用いることや, 対義語を用いても良い(なお, 対義語の場合はマイナスをつけて$-\bm{v}^{T}\bm{q}$とする必要がある).
  その他にも, 現在画像にノイズを加えて平均を取ったり, 複数のモデルを利用したりすることも可能である.
  \cite{kanazawa2023cooking}では, 対立する2つのテキストを用意した場合について実験を行っている.
}%

\subsection{Robotic Continuous State Recognition Using Black-Box Optimization} \label{subsec:bb-optimization}
\begin{figure}[t]
  \centering
  \includegraphics[width=0.75\columnwidth]{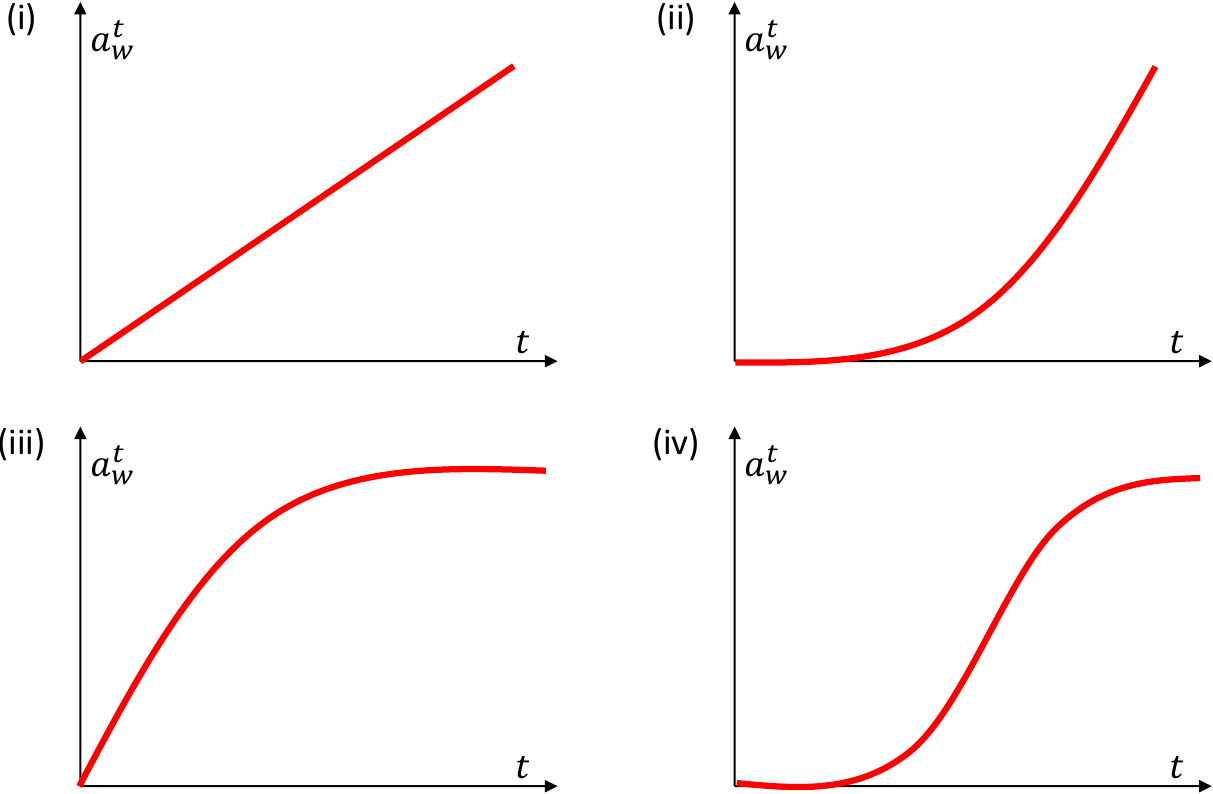}
  \vspace{-1.0ex}
  \caption{Types of continuous state changes. All of these changes can be represented by a sigmoid function.}
  \label{figure:func-shape}
  \vspace{-2.0ex}
\end{figure}

\switchlanguage%
{%
  There are several challenges with the previously described method.
  First, preliminary experiments show that the recognition performance varies greatly depending on the choice of texts.
  By using a variety of texts, we can absorb the differences in recognition performance among texts and obtain stable recognition performance.
  However, if diverse texts are used uniformly, texts with low recognition performance may have a negative impact.
  In addition, the recognition of the beginning and end of state changes relies on human thresholding.
  It is desirable to have a system that automatically obtains high-performance state recognition with as little human intervention as possible.

  Therefore, we propose a method to automatically obtain a high-performance state recognizer by adjusting the weighting of various texts, as shown in \figref{figure:proposed}.
  We obtain data on the state changes only once, fit the continuous similarity changes to a sigmoid function, compute an evaluation function, and adjust the weighting of each text based on black-box optimization.
  By obtaining similarity changes with larger state changes and smaller variance, accurate and robust continuous state recognition can be achieved.
  Here, no annotation of the data is required.
  Moreover, there is no need to prepare a model or program for each state recognition, as only the text set and weighting need to be changed for each state recognition, which facilitates the management of source codes and computational resources.

  Here, we discuss the reason for using the sigmoid function for fitting.
  As shown in \figref{figure:func-shape}, there are basically four possible state changes: (i), (ii), (iii), and (iv) (note that the vertical axis of the graph, $a^{t}_{w}$, is the weighted similarity of \equref{eq:similarity} described subsequently).
  They are (i) the case where the state change continues at all times, (ii) the case where there is no change at the beginning but a continuous change happens thereafter, (iii) the case where a state change occurs from the beginning but the change eventually converges, and (iv) the case where (ii) and (iii) are combined.
  The sigmoid function can represent all of these cases by changing the slope of its shape or by shifting it to the left or right, and is suitable for representing continuous state changes.
  It is also useful in that the value range falls within $(0, 1)$, which allows automatic threshold design such that the state change is considered to have ended when, for example, the value reaches the 80\% change point (0.8).
}%
{%
  先の方法では, いくつかの問題がある.
  まず, 予備実験から, 言語テキストの選び方によって認識性能が大きく変化することがわかっている.
  多様なテキストを用いることで, テキストごとの認識性能の差を吸収し, 安定した認識性能を得ることができる.
  しかし, 多様なテキストを一様に使ってしまうと, 認識性能の低いテキストが悪影響を及ぼす可能性がある.
  また, 状態変化の始まりや終了の認識を, 人間による閾値設定に頼っている.
  できるだけ人間の介入を減らし, 性能の高い状態認識を自動的に得る仕組みが望ましい.

  そこで本研究では, 多様なテキストを用意した上で, それらの重みづけを調整することで, 性能の高い状態認識器を自動的に得る手法を提案する.
  一度だけ状態変化に関するデータを取得し, その際の連続的な類似度変化をsigmoid関数にフィッティング, 評価関数を計算し, ブラックボックス最適化に基づき各テキストの重みづけを調整する.
  より状態変化量が大きく, 分散の小さな類似度変化を得ることで, 正確でロバストな連続状態認識が可能になる.
  なお, この際データについて一切のアノテーション等は必要ない.
  また, 状態認識ごとに変化させるのはテキスト集合と重み付けのみで良く, 各状態認識ごとにモデルやプログラムを用意する必要もない.

  ここで, フィッティングにsigmoid関数を用いる理由について述べておく.
  \figref{figure:func-shape}に示すように, 基本的に状態変化は(i), (ii), (iii), (iv)の4種類があり得る(なお, グラフの縦軸である$a^{t}_{w}$は\equref{eq:similarity}で示す重み付けした類似度である).
  これは, (i)状態変化が常に継続する場合, (ii)最初は変化がなくその後継続的に変化する場合, (iii)最初から変化が起こるが最終的に変化が収束する場合, (iv)最初は変化がなくその後変化が起こり最終的に変化が収束する場合である.
  sigmoid関数はその形状の傾きを変化させたり, 左右にずらしたりすることで, これらを全て表現することができ, 連続的な状態変化を表すのに適している.
  また, 値域が$(0, 1)$の間に収まるため, 例えば80\%変化点(0.8)になった際に状態変化が終了したと見なす, というような自動的な閾値設計が可能になる点でも有用である.
}%

\begin{figure}[t]
  \centering
  \includegraphics[width=0.75\columnwidth]{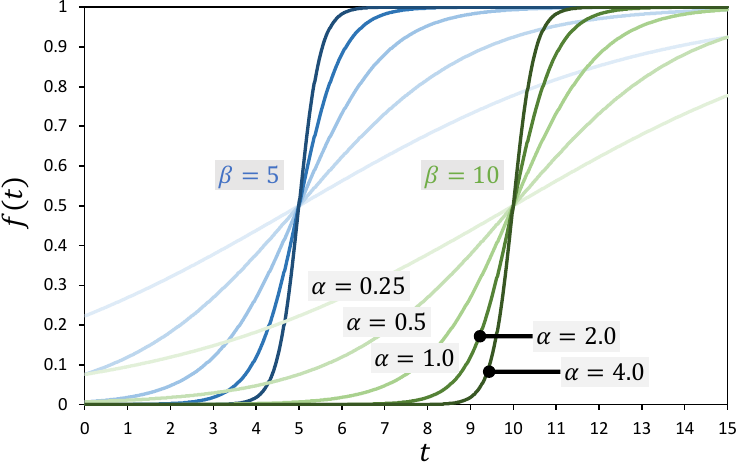}
  \vspace{-1.0ex}
  \caption{Changes in the sigmoid function when changing the parameters $\alpha$ and $\beta$. By increasing $\alpha$, the sigmoid function becomes steeper, which means that the state change can be detected more easily. By increasing $\beta$, the state change is less likely to be misidentified early in the process.}
  \label{figure:sigmoid-alpha}
  \vspace{-2.0ex}
\end{figure}

\switchlanguage%
{%
  In the following, we describe the optimization procedure.
  First, the data $D$ of the state change to be recognized is obtained once.
  This $D$ consists of a time series of images $V_{t}$ ($1 \leq t \leq T$, where $T$ denotes the number of images).
  We also prepare a set of texts $Q_i$ ($1 \leq i \leq N$) that describe the state change to be recognized.
  Here, let $Q^{1}_{i}$ (e.g. ``boiled water'' and ``melted butter'') be the set of texts that indicate the change has occurred, and $Q^{-1}_{i}$ (e.g. ``unboiled water'' and ``unmelted butter'') be the set of texts that indicate the change has not occurred.

  Next, for each weight $w_{i}$ ($1 \leq i \leq N$, $0 \leq w_i \leq 1$) of each text, we set an evaluation function $E$ to be maximized based on black-box optimization.
  Here, $w_{i}$ represents the importance of each text and how to weigh recognition results computed from each text.
  First, given a weight $\bm{w}$, the similarity $a^{t}_{w}$ between the current image $V_t$ and the text set $Q$ is calculated as follows,
  \begin{align}
    a^{t}_{w} := \sum^{N}_{i}{p_{i}w_{i}\bm{v}_{t}^{T}\bm{q}_{i}} / \sum^{N}_{i}{w_i} \label{eq:similarity}
  \end{align}
  where $p_{i}$ is a variable that returns $1$ for $Q^{1}_{i}$ and $-1$ for $Q^{-1}_{i}$.
  It can be said that the similarity $\bm{v}^{T}_{t}\bm{q}_{i}$ for each text is weighted by $p_{i}w_{i}$.
  The continuous change of this value is fitted to a sigmoid function.
  The sigmoid considered in this study has the following form,
  \begin{align}
    f(t) := \frac{1}{1 + e^{-\alpha (t-\beta)}}
  \end{align}
  where $\alpha$ and $\beta$ are adjustable parameters that determine the shape of the sigmoid.
  As shown in \figref{figure:sigmoid-alpha}, the larger the $\alpha$ is, the larger the slope of the sigmoid function becomes, and the more clearly the similarity changes.
  $\beta$ is a parameter that shifts the center of the sigmoid function along the $t$-axis, and the larger it is, the less likely the state change is misidentified early in the process, since the change in $f(t)$ occurs at the end of process.
  Since $0 < f(t) < 1$ for this sigmoid function, the scale of $a^{t}_{w}$ must be changed.
  In this study, we compute $\hat{a}^{t}_{w}$ where $a^{t}_{w}$ is scaled as follows,
  \begin{align}
    \hat{a}^{t}_{w} := \frac{a^{t}_{w}-a^{min}_{w}}{a^{max}_{w}-a^{min}_{w}} \label{eq:scale}
  \end{align}
  where $a^{\{min, max\}}_{w}$ denotes the minimum and maximum values in the moving average of $a^{t}_{w}$ ($1 \leq t \leq T$) over 3 seconds.
  This makes $0 \leq \hat{a}^{t}_{w} \leq 1$, which facilitates fitting to $f(t)$.
  During inference, it is important to note that the entire time series data $D$ is not available from the beginning, and so \equref{eq:scale} is performed using the computed $a^{\{min, max\}}_{w}$ during the optimization process.
  Assuming that the default parameter of $(\alpha, \beta)$ is $(0.1, T/2)$, we obtain $(\alpha, \beta)$ by fitting using the nonlinear least-squares method.
  Note that the constraints $\alpha \geq 0$ and $\beta \geq 0$ are imposed in the fitting.
  From these, we define the evaluation function $E$ to be maximized as follows,
  \begin{align}
    E(\bm{w}) := \alpha \beta / \sigma
  \end{align}
  where $\sigma$ is the root mean squared error of the fitting.
  In other words, the evaluation function is designed to minimize the fitting error while making the amount of change as large as possible and ensuring that a large change in $f(t)$ occurs at the end of the state change as possible.

  Finally, we perform black-box optimization.
  In this study, we apply a genetic algorithm using the library DEAP \cite{fortin2012deap} as the algorithm.
  The gene sequence represented by $w_i$ is optimized based on the maximization of $E$.
  The function cxBlend is used for crossover with a probability of 50\%, and mutGaussian is used for mutation with a probability of 20\% with mean 0 and variance 0.1.
  Individuals are selected by the function selTournament, where the tournament size is set to 5, the number of individuals is set to 300, and the number of generations is set to 300.
  The choice of optimization method is flexible, and we have tried several methods such as Tree-structured Parzen Estimator (TPE) and Covariance Matrix Adaptation Evolution Strategy (CMAES), but we did not observe significant differences in the results.
  Therefore, we used a common genetic algorithm with minimum computational cost.
}%
{%
  以降では最適化の手順について述べる.
  まず, 認識したい状態変化に関するデータ$D$を一度取得する.
  この$D$は時系列の画像$V_{t}$ ($1 \leq t \leq T$, $T$は画像の枚数を表す)によって構成される.
  また, 認識したい状態変化に関するテキスト集合$Q_i$ ($1 \leq i \leq N$)を用意する.
  このとき, 変化したことを表すテキスト集合を$Q^{1}_{i}$ (e.g. ``boiled water'' and ``melted butter''), 変化していないことを表すテキスト集合を$Q^{-1}_{i}$ (e.g. ``unboiled water'' and ``unmelted butter'')とする.

  次に, 各テキストに対する重み$w_{i}$ ($1 \leq i \leq N$, $0 \leq w_i \leq 1$)について, ブラックボックス最適化に基づいて最大化する評価関数$E$を設定する.
  まず, 重み$\bm{w}$が与えられたとき, 現在の画像$V_t$とテキスト集合$Q$との間の類似度$a^{t}_{w}$は以下のように計算される.
  \begin{align}
    a^{t}_{w} := \sum^{N}_{i}{p_{i}w_{i}\bm{v}_{t}^{T}\bm{q}_{i}} / \sum^{N}_{i}{w_i} \label{eq:similarity}
  \end{align}
  ここで, $p_{i}$は, $Q^{1}_{i}$に対して$1$を, $Q^{-1}_{i}$に対して$-1$を返す変数である.
  各テキストにおける類似度$\bm{v}^{T}_{t}\bm{q}_{i}$を$p_{i}w_{i}$によって重み付けしたものと言える.
  この値の連続的な変化をsigmoid関数にフィッティングする.
  本研究で考えるsigmoidは以下のような形とする.
  \begin{align}
    f(t) := \frac{1}{1 + e^{-\alpha (t-\beta)}}
  \end{align}
  ここで, $\alpha$と$\beta$はsigmoidの形状を決定する調整可能なパラメータである.
  \figref{figure:sigmoid-alpha}に示すように, $\alpha$が大きいほどシグモイド関数の傾きが大きくなり状態変化量がはっきりとわかりやすい関数になる.
  $\beta$はシグモイド関数の中心を$t$軸方向にずらすパラメータであり, より大きいほど, $f(t)$の変化が時間的に最後の方に起こるため, 前の方の時間で状態変化を誤認識しにくくなる.
  このシグモイド関数は$0 < f(t) < 1$であるため, $a^{t}_{w}$のスケールを変化させる必要がある.
  本研究では, 以下のようにスケールした$\hat{a}^{t}_{w}$を計算する.
  \begin{align}
    \hat{a}^{t}_{w} := \frac{a^{t}_{w}-a^{min}_{w}}{a^{max}_{w}-a^{min}_{w}} \label{eq:scale}
  \end{align}
  ここで, $a^{\{min, max\}}_{w}$は$a^{t}_{w}$ ($1 \leq t \leq T$)の3秒間に渡る移動平均の最小値と最大値を表す.
  これにより, $0 \leq \hat{a}^{t}_{w} \leq 1$となるため, $f(t)$へのフィッティングが容易になる.
  なお, 推論時は最初から全体の時系列データ$D$が得られるわけではないため, 最適化時に計算された$a^{\{min, max\}}_{w}$を用いて\equref{eq:scale}の変換を行う.
  $(\alpha, \beta)$のデフォルトパラメータを$(0.1, T/2)$として, 非線形最小二乗法によってフィッティングを行い$(\alpha, \beta)$を求める.
  なお, フィッティングの際には, $\alpha\geq0$, $\beta\geq0$という制約を課す.
  これらから, 最大化する評価関数$E$を以下のように定義する.
  \begin{align}
    E(\bm{w}) := \alpha \beta / \sigma
  \end{align}
  ここで, $\sigma$はフィッティングの誤差の平均二乗平方根を表す.
  つまり, フィッティングへの誤差を小さくしつつ, なるべく変化量を大きくし, かつなるべく$f(t)$の大きな変化が状態変化終盤に起こるようにする評価関数と言える.

  最後に, ブラックボックス最適化を行う.
  本研究ではアルゴリズムとして, DEAP \cite{fortin2012deap}を用いた遺伝的アルゴリズムを適用する.
  $w_i$により表現された遺伝子配列を, $E$の最大化に基づき最適化する.
  この際, 50\%の確率で関数cxBlendにより交叉, 20\%の確率で平均0, 分散0.1の関数mutGaussianにより突然変異を行う.
  個体選択は関数selTournamentで行い, トーナメントサイズを5, 個体数を300, 世代数を300とする.
  なお, 最適化手法は何でも良く, Tree-structured Parzen Estimator (TPE)やCovariance Matrix Adaptation Evolution Strategy (CMAES)なども試しましたが, 結果に大きな差を得られなかったため, 最も計算量の低い遺伝的アルゴリズムを使用しています.
}%

\begin{figure*}[t]
  \centering
  \includegraphics[width=1.9\columnwidth]{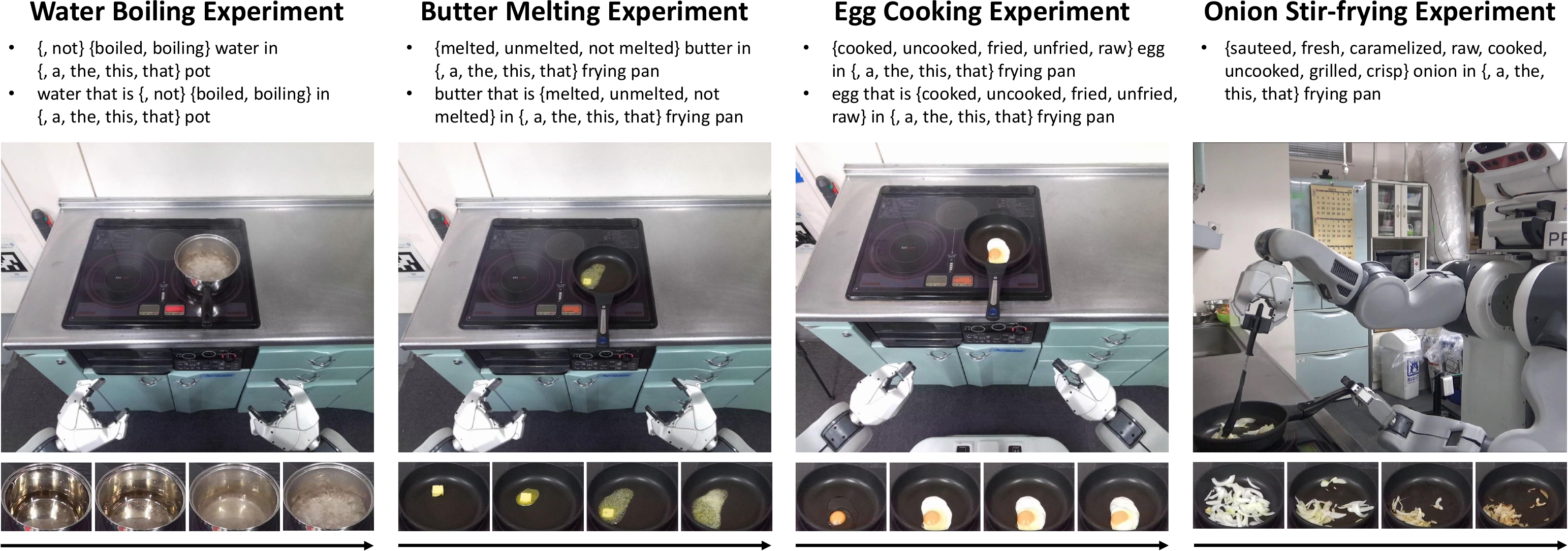}
  \vspace{-1.0ex}
  \caption{The experimental setup: the text prompts and representative images for water boiling, butter melting, egg cooking, and onion stir-frying experiments.}
  \label{figure:exp-setup}
  \vspace{-1.0ex}
\end{figure*}

\section{Experiments} \label{sec:experiment}
\switchlanguage%
{%
  In this study, we perform four experiments: recognition of water boiling, butter melting, egg cooking (fried egg), and onion stir-frying.
  Each experiment with the prepared text set $Q$ are shown in \figref{figure:exp-setup}.
  First, we prepare a dataset $D_{opt}$ for optimization and a dataset $D_{eval}$ for evaluation.
  The state changes of water, butter, and egg are obtained at 10 Hz when the mobile robot PR2 looks at the stove that is heated to a certain degree.
  For onion stir-frying, the image of the onion is acquired every time PR2 stirs the pan with a spatula, because the onion would get burned if not stirred constantly.
  Although all experiments are conducted at the same heat intensity, the length $T$ of each data is slightly different.
  For each experiment, the time when the state change ends is annotated as $t_{data}$.
  Next, we prepare at most a set of 50 texts $Q$ describing the state change for each experiment.
  We prepare a large number of $Q$ by changing the article, state expression, and expression form.
  We use five kinds of articles: ``a'', ``the'', ``this'', ``that'', and no article.
  For state expressions, antonyms such as ``boiled''/``unboiled'' and ``melted''/``not melted'' are used (synonyms are also used).
  The expression form is slightly changed, such as ``boiled water'' and ``water that is boiled'' are used.

  In this study, we conduct experiments using two models, CLIP and ImageBind.
  For each model, we evaluate both $D_{opt}$ and $D_{eval}$ datasets in three settings, \textbf{OPT}, \textbf{ONE}, and \textbf{ALL}.
  \textbf{OPT} is the result of applying the black-box optimization proposed in this study.
  \textbf{ONE} is the result when only the best $Q$ that maximizes $E$ is used among the prepared $Q$.
  Using only one best $Q$ means that a state in which only one of the $N$ scalar values in $\bm{w}$ is 1 and the rest are 0 is created, and the $\bm{w}$ with the highest $E$ is selected.
  \textbf{ALL} is the result when all the prepared $Q$ are used equally without optimization.
  This means that $w_i=1$ ($1 \leq i \leq N$).
  Regarding $D_{opt}$, we plot the transition of $\hat{a}^{t}_{w}$ and its moving average over 3 seconds for \textbf{OPT}, \textbf{ONE}, and \textbf{ALL}, respectively.
  Note that the moving averages are not plotted for the onion stir-frying experiment, since the number of images is small.
  Regarding $D_{eval}$, we plot $\hat{a}^{t}_{w}$, which is transformed by each \equref{eq:scale} obtained from \textbf{OPT}, \textbf{ONE}, and \textbf{ALL} in $D_{opt}$, and its moving average.
  For each plot, $t_{detected}$ is the time when the moving average first exceeds the set threshold $C_{thre}$ ($C_{thre}=0.8$ in this study).
  As $t_{diff}=|t_{detected}-t_{data}|$, $t_{diff}$ should be as small as possible.
  For each experiment, the evaluation value $E$ when fitting the change in $\hat{a}^{t}_{w}$ into the sigmoid function $f(t)$ is described.
  Note that while $E$ is appropriate regarding $D_{opt}$, it is for reference only regarding $D_{eval}$, since $\hat{a}^{t}_{w}$ may not fall between [0, 1] depending on the experiment.

  Finally, as a cooking experiment utilizing the proposed method, the PR2 robot boils water, blanches broccoli, and stir-fries it with melted butter.
}%
{%
  本研究では基礎的な実験として, 水の沸騰認識, バターの解け具合認識, 卵の固まり具合認識, 玉ねぎの炒まり具合認識の4つの実験を行う.
  各実験の様子と用意したテキスト集合$Q$を\figref{figure:exp-setup}に示す.
  まず, 最適化用のデータセット$D_{opt}$と評価用のデータセット$D_{eval}$を用意する.
  台車型ロボットPR2がコンロに視線を向け, ある一定の火力を加えた際の, 水・バター・卵の状態変化を10Hzで取得している.
  玉ねぎについては, 動かさずに置いておくと焦げるため, PR2がヘラで混ぜる動作を行う度に画像を取得している.
  同じ火力で実験しているが, それぞれデータの長さ$T$は多少異なる.
  各実験について, 状態変化の終了時刻を$t_{data}$としてアノテーションしている.
  次に, 各実験について, 状態変化を記述するテキスト集合$Q$を最大で50個用意する.
  ここで, 冠詞・状態表現・表現形式を変更することで多数の$Q$を用意している.
  冠詞については, ``a'', ``the'', ``this'', ``that'', 無冠詞の5種類を用いる.
  状態表現については, ``boiled''や``unboiled'', ``melted''や``not melted''のような対義語を用いる(類義語の使用も行う).
  表現形式については, ``boiled water''や``water that is boiled''のような表現形式の変化を用いる.

  比較実験について述べる.
  本研究では, CLIPとImageBindの2つのモデルを用いて実験を行う.
  また, それぞれのモデルについて, \textbf{OPT}, \textbf{ONE}, \textbf{ALL}の3つの設定で, $D_{opt}$と$D_{eval}$の両者のデータセットについて評価する.
  \textbf{OPT}は本研究におけるブラックボックス最適化を適用した際の結果である.
  \textbf{ONE}は, 用意した$Q$の中から$E$を最大化する最善の$Q$を一つだけ選択した際の結果である.
  Using only one $Q$, that is, creating a state in which only one of the $N$ scalar values in $\bm{w}$ is 1 and the rest are 0, the $\bm{w}$ with the highest $E$ among these created $\bm{w}$ is selected.
  \textbf{ALL}は, 最適化を用いず, 用意した$Q$を全て均等に使った際の結果である.
  $D_{opt}$について, \textbf{OPT}, \textbf{ONE}, \textbf{ALL}, それぞれにおける$\hat{a}^{t}_{w}$の遷移とその3秒ごとの移動平均をプロットする.
  なお, 玉ねぎについては画像の枚数が非常に少ないため, 移動平均は行わない.
  $D_{eval}$については, $D_{opt}$において\textbf{OPT}, \textbf{ONE}, \textbf{ALL}, それぞれで得られた\equref{eq:scale}と同じ変換を行った$\hat{a}^{t}_{w}$とその移動平均を同様にプロットする.
  それぞれのプロットについて, 移動平均が設定した閾値$C_{thre}$を初めて越した時刻を$t_{detected}$とする(本研究では$C_{thre}=0.8$とする).
  $t_{diff}=|t_{detected}-t_{data}|$として, $t_{diff}$がより小さいことが望ましい.
  また, 各実験について, シグモイド関数$f(t)$へのフィッティングを行った際の評価値$E$を記載している.
  なお, この値は$D_{opt}$については適切であるが, $D_{eval}$については実験次第で$\hat{a}^{t}_{w}$が[0, 1]の間に収まらない場合があるため, $E$はあくまで参考程度の値である.

  最後に, 本手法を利用した調理実験として, 湯を沸かし, ブロッコリーを茹で, 溶かしたバターと醤油で炒める実験を行う.
}%

\begin{figure*}[t]
  \centering
  \includegraphics[width=1.95\columnwidth]{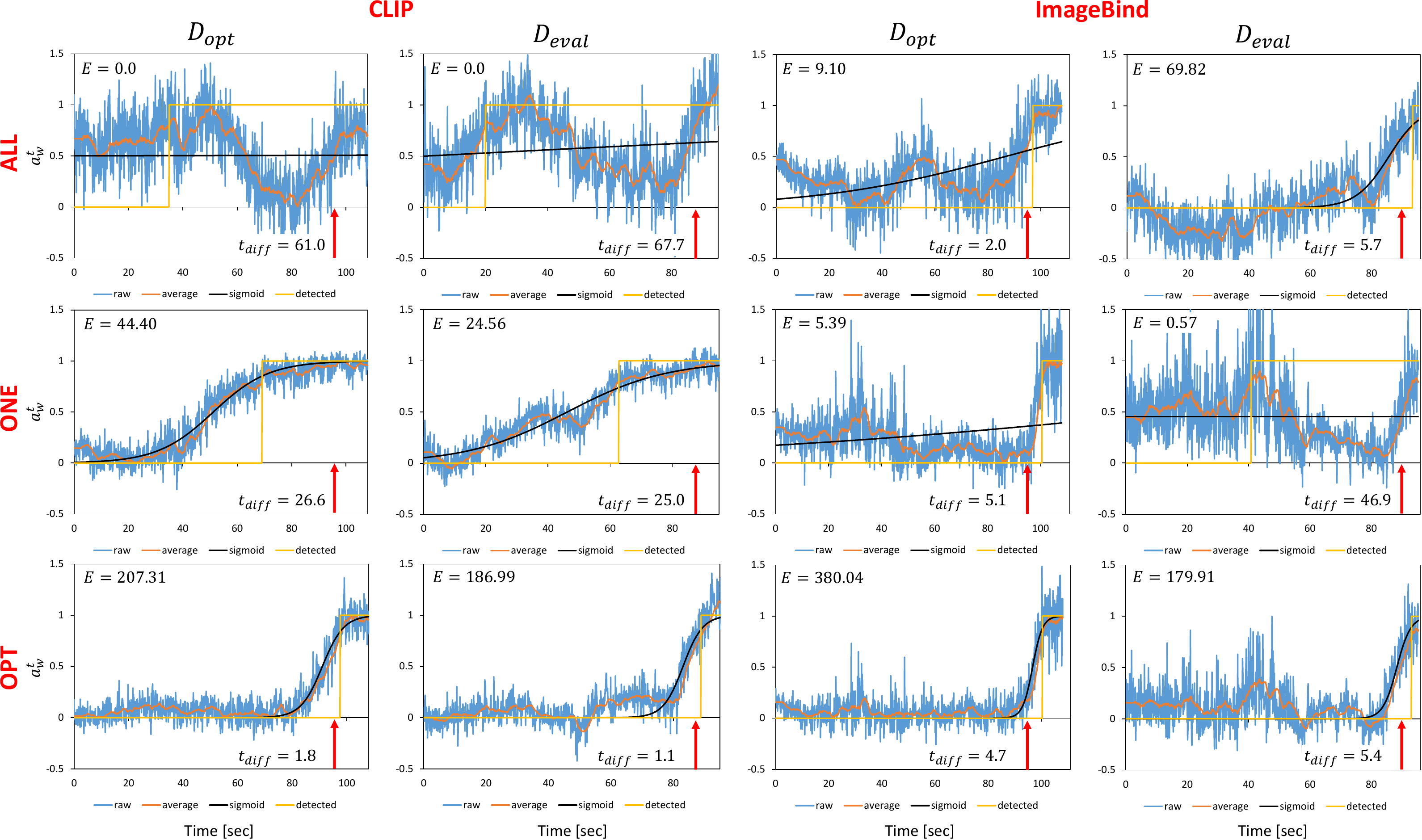}
  \vspace{-1.0ex}
  \caption{Results of the water boiling experiment. For the two models CLIP and ImageBind, the results of \textbf{OPT}, \textbf{ONE}, and \textbf{ALL} are shown regarding $D_{opt}$ and $D_{eval}$. In the graphs, ``raw'' expresses the raw value of similarity, ``average'' expresses the moving average of the raw value over 3 seconds, ``sigmoid'' expresses the sigmoid function fitted to ``average'', and ``detected'' expresses the function that becomes 1 after $t_{detected}$. The red arrow shows $t_{data}$, the annotated time of state change.}
  \label{figure:water-exp}
  \vspace{-2.0ex}
\end{figure*}

\subsection{Water Boiling Experiment} \label{subsec:water-exp}
\switchlanguage%
{%
  The results of the water boiling experiment are shown in \figref{figure:water-exp}.
  Here, ``raw'' is the raw value, ``average'' is the moving average, ``sigmoid'' is the result of fitting $f(t)$ to $\hat{a}^{t}_{w}$, ``detected'' is the function that becomes 1 after $t_{detected}$, and the red arrow indicates $t_{data}$.
  As for CLIP, the change in similarity of \textbf{ALL} fluctuates and does not change proportionally with the boiling state.
  Thus, the fitting is not successful, the evaluation value $E$ is zero, and $t_{diff}$ is large.
  On the other hand, the change in similarity of \textbf{ONE} gradually increases with time, thus $E$ is larger and $t_{diff}$ is smaller than that of \textbf{ALL}.
  However, because of the gradual change in the similarity, the state is determined to be boiling earlier than in actuality.
  In contrast, for \textbf{OPT}, the abrupt change in similarity occurs almost simultaneously with boiling, thus $E$ is the largest and $t_{diff}$ is quite small (about 1 second).
  The variance of the similarity changes is also small and stable recognition results are obtained.
  As for ImageBind, reasonable performance is obtained even for \textbf{ALL} and \textbf{ONE}, and $t_{diff}$ is relatively small.
  The change in similarity of \textbf{OPT} is more stable than that of \textbf{ALL} and \textbf{ONE}, indicating higher performance.
  Note that the top 5 text prompts and their weights are \textit{water that is not boiling in the pot} (0.12), \textit{water that is boiled in pot} (0.12), \textit{water that is not boiling in the pot} (0.12), \textit{water that is not boiling in this pot} (0.12), and \textit{boiling water in that pot} (0.11) (the weight is normalized to be $\Sigma{w_i}=1$).
}%
{%
  水の沸騰認識実験を行った結果を\figref{figure:water-exp}に示す.
  ここで, ``raw''は生値, ``average''は移動平均, ``sigmoid''は$f(t)$を$\hat{a}^{t}_{w}$にフィッティングした結果, ``detected''は$t_{detected}$において1となる関数, 赤い矢印は$t_{data}$を示している.
  まずCLIPについて, \textbf{ALL}の場合は大きく値が波打っており, 沸騰とともに値の動きが比例して変化していない.
  実際, フィッティングに成功しておらず, 評価関数$E=0$であり, $t_{diff}$も非常に大きい.
  一方\textbf{ONE}では, 時間が進むに連れて徐々に類似度が上昇しており, \textbf{ALL}に比べて$E$は大きく, $t_{diff}$は小さくなっている.
  しかし, なだらかに徐々に類似度が変化するため, 実際に沸騰した時間よりもかなり前にに沸騰と判定してしまっている.
  これに対して\textbf{OPT}では, 類似度の急激な変化が沸騰とほぼ同時に起こっており, $E$は最も大きく, $t_{diff}$は1秒程度とかなり小さい.
  また, 生値の分散が小さく, 安定した認識結果が得られている.
  ImageBindについては, \textbf{ALL}や\textbf{ONE}についてもそれなりの性能が得られており, $t_{diff}$は比較的小さい.
  \textbf{OPT}については, \textbf{ALL}や\textbf{ONE}に比べて生値が安定しており, より高い性能を示している.
  なお, 最適化によって得られた上位5つのテキストプロンプトと重みは, \textit{water that is not boiling in the pot} (0.12), \textit{water that is boiled in pot} (0.12), \textit{water that is not boiling in the pot} (0.12), \textit{water that is not boiling in this pot} (0.12), and \textit{boiling water in that pot} (0.11)であった($\Sigma{w_i}=1$となるように正規化している).
}%

\begin{figure*}[t]
  \centering
  \includegraphics[width=1.95\columnwidth]{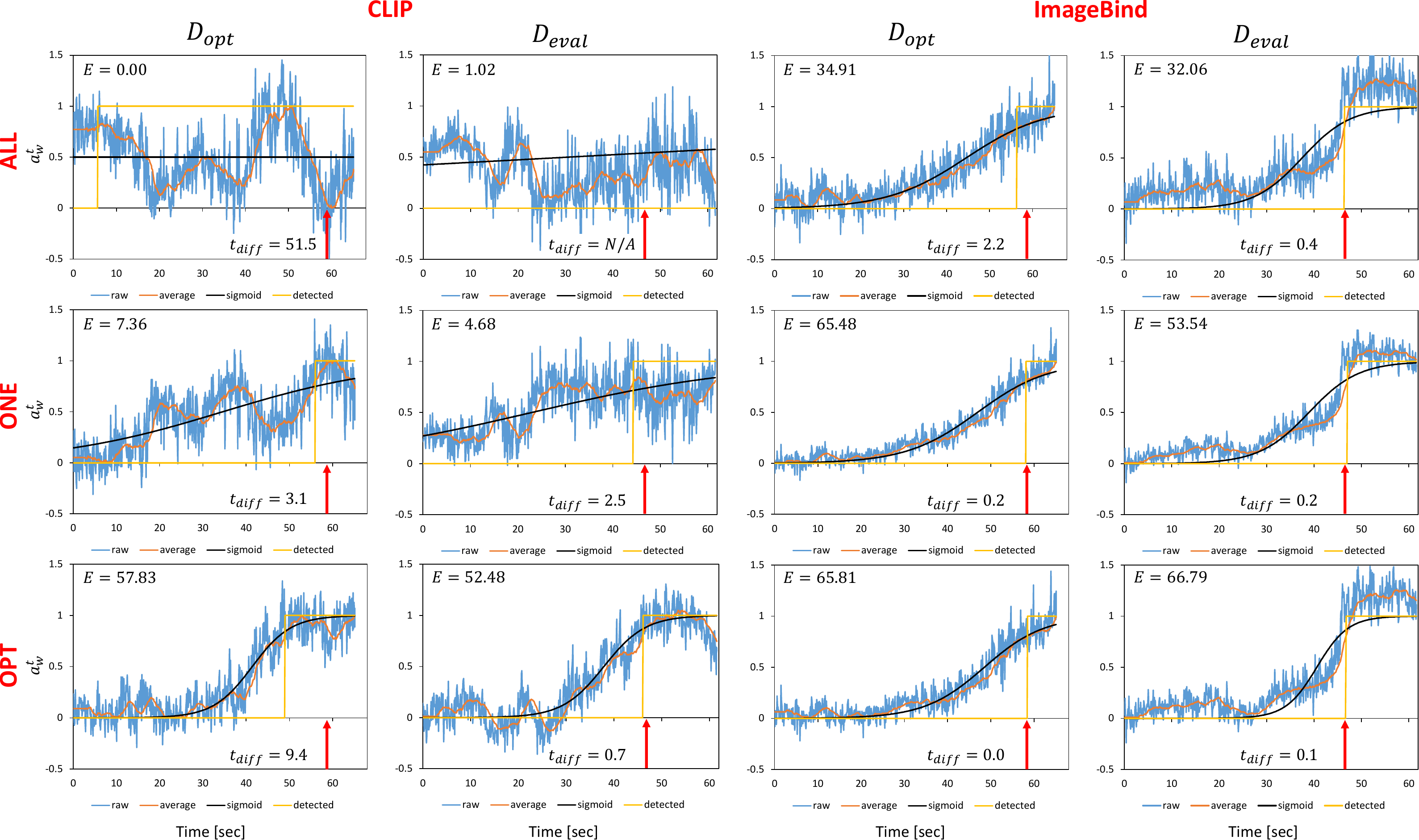}
  \vspace{-1.0ex}
  \caption{Results of the butter melting experiment. For the two models CLIP and ImageBind, the results of \textbf{OPT}, \textbf{ONE}, and \textbf{ALL} are shown regarding $D_{opt}$ and $D_{eval}$. In the graphs, ``raw'' expresses the raw value of similarity, ``average'' expresses the moving average of the raw value over 3 seconds, ``sigmoid'' expresses the sigmoid function fitted to ``average'', and ``detected'' expresses the function that becomes 1 after $t_{detected}$. The red arrow shows $t_{data}$.}
  \label{figure:butter-exp}
  \vspace{-2.0ex}
\end{figure*}

\subsection{Butter Melting Experiment} \label{subsec:butter-exp}
\switchlanguage%
{%
  The results of the butter melting experiment are shown in \figref{figure:butter-exp}.
  As for CLIP, the change in similarity of \textbf{ALL} fluctuates as with \secref{subsec:water-exp}, and does not change proportionally with the degree of melting.
  Thus, $E$ is small and $t_{diff}$ is large.
  The performance of \textbf{ONE} is better than that of \textbf{ALL}, but the change in similarity still fluctuates.
  Compared to \textbf{ONE}, \textbf{OPT} shows a stable change in the similarity, but the state change is detected earlier, especially for $D_{opt}$.
  As for ImageBind, state changes are recognized with high accuracy for all settings, thus $E$ is large and $t_{diff}$ is small.
  \textbf{ALL} has a larger variance of similarity changes compared to \textbf{ONE} and \textbf{OPT}, resulting in $E$ being approximately halved.
  Note that the top 5 text prompts and their weights are \textit{butter that is not melted in that frying pan} (0.24), \textit{butter that is not melted in frying pan} (0.24), \textit{not melted butter in a frying pan} (0.19), \textit{not melted butter in that frying pan} (0.18), and \textit{melted butter in frying pan} (0.08).
}%
{%
  バターの溶け具合認識実験を行った結果を\figref{figure:butter-exp}に示す.
  まずCLIPについて, \textbf{ALL}の場合は\secref{subsec:water-exp}と同様に大きく値が波打っており, 溶け具合とともに値の動きが比例して変化していない.
  $E$は小さく$t_{diff}$は大きく, $D_{eval}$についてはそもそも状態変化が検知されなかった.
  \textbf{ONE}は\textbf{ALL}に比べると性能は良いが, 類似度は大きく波打っている.
  \textbf{OPT}は\textbf{ONE}に比べ類似度の変化が安定しているが, 特に$D_{opt}$では早めに状態変化が検知されてしまっている.
  ImageBindについては, \textbf{ALL}, \textbf{ONE}, \textbf{OPT}の全てにおいて, 非常に高い精度で認識が出来ており, $E$は大きく$t_{diff}$は非常に小さい.
  \textbf{ALL}は, \textbf{ONE}や\textbf{OPT}に比べると類似度の分散が大きいため, $E$は半分程度となっている.
}%
\begin{figure*}[t]
  \centering
  \includegraphics[width=1.95\columnwidth]{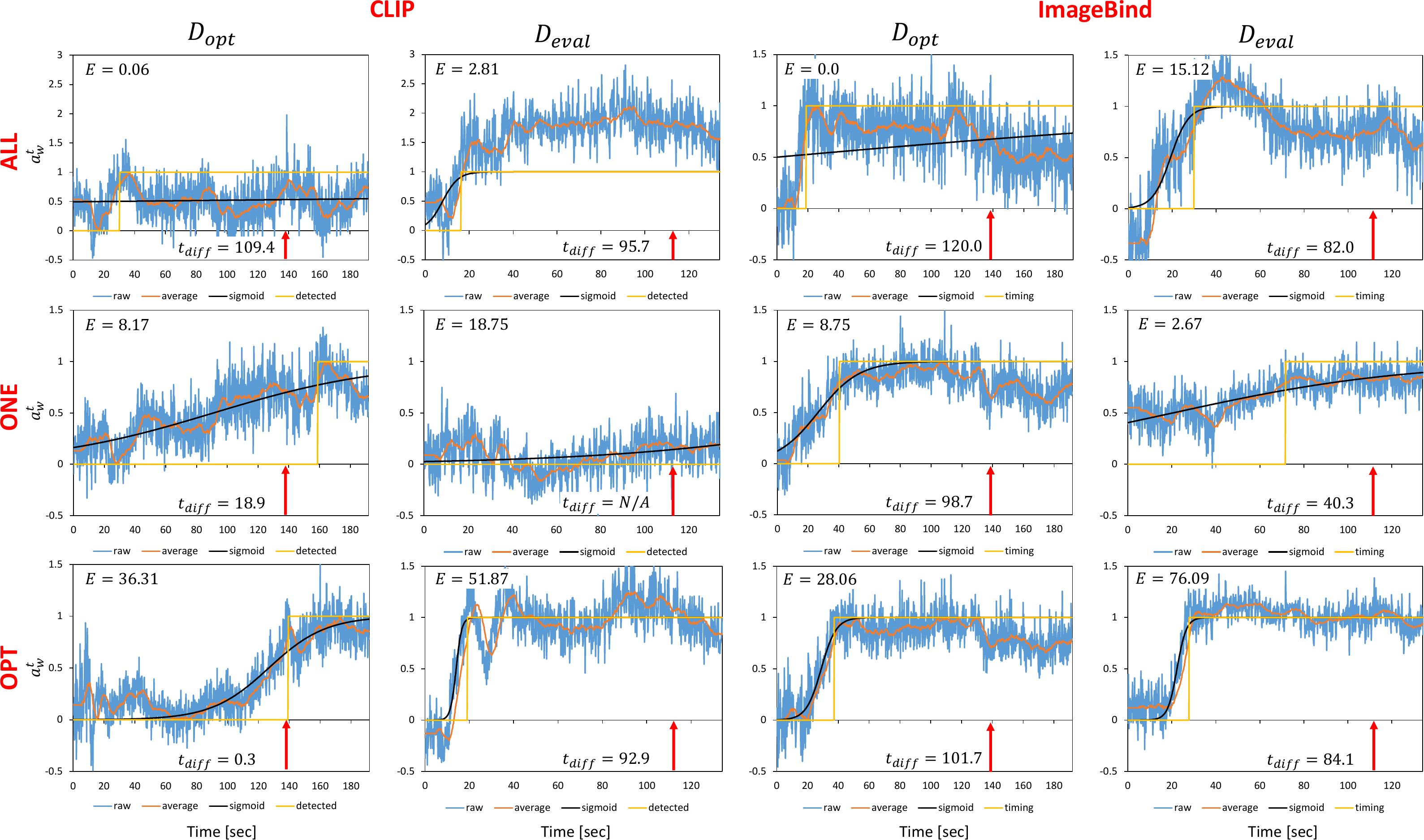}
  \vspace{-1.0ex}
  \caption{Results of the egg cooking experiment. For the two models CLIP and ImageBind, the results of \textbf{OPT}, \textbf{ONE}, and \textbf{ALL} are shown regarding $D_{opt}$ and $D_{eval}$. In the graphs, ``raw'' expresses the raw value of similarity, ``average'' expresses the moving average of the raw value over 3 seconds, ``sigmoid'' expresses the sigmoid function fitted to ``average'', and ``detected'' expresses the function that becomes 1 after $t_{detected}$. The red arrow shows $t_{data}$.}
  \label{figure:egg-exp}
  \vspace{-3.0ex}
\end{figure*}

\begin{figure*}[t]
  \centering
  \includegraphics[width=1.95\columnwidth]{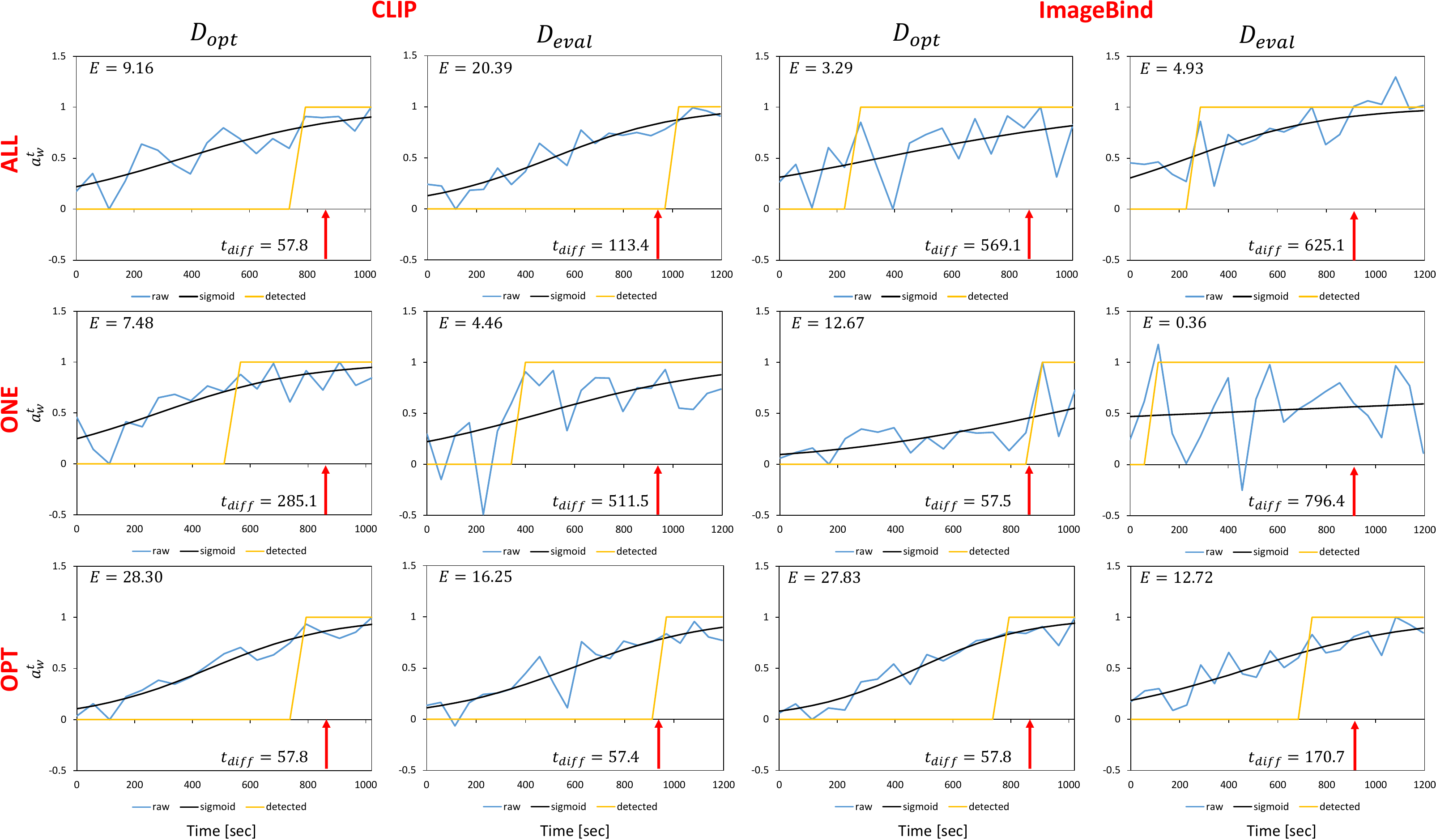}
  \vspace{-1.0ex}
  \caption{Results of the onion stir-frying experiment. For the two models CLIP and ImageBind, the results of \textbf{OPT}, \textbf{ONE}, and \textbf{ALL} are shown regarding $D_{opt}$ and $D_{eval}$. In the graphs, ``raw'' expresses the raw value of similarity, ``sigmoid'' expresses the sigmoid function fitted to ``raw'', and ``detected'' expresses the function that becomes 1 after $t_{detected}$. The red arrow shows $t_{data}$.}
  \label{figure:onion-exp}
  \vspace{-2.0ex}
\end{figure*}

\subsection{Egg Cooking Experiment} \label{subsec:egg-exp}
\switchlanguage%
{%
  The results of the egg cooking experiment are shown in \figref{figure:egg-exp}.
  As for CLIP, the change in similarity of \textbf{ALL} fluctuates as with \secref{subsec:water-exp} and \secref{subsec:butter-exp}, and thus $E$ is small and $t_{diff}$ is large.
  For $D_{opt}$, the performance of \textbf{ONE} and \textbf{OPT} is reasonable and $t_{diff}$ is small.
  On the other hand, the results for $D_{eval}$ are significantly different from those for $D_{opt}$, and the accuracy is low.
  As for ImageBind, there is no large difference between $D_{opt}$ and $D_{eval}$ as with CLIP, but in most cases, the similarity increases significantly in the early phase of the state change and then remains constant.
  Therefore, the state change cannot be detected properly, thus $t_{diff}$ is large.
  Note that the top 5 text prompts and their weights are \textit{raw egg in that frying pan} (0.07), \textit{cooked egg in a frying pan} (0.07), \textit{cooked egg in the frying pan} (0.07), \textit{cooked egg in this frying pan} (0.07), and \textit{egg that is fried in frying pan} (0.06).
}%
{%
  目玉焼きの固まり具合認識実験を行った結果を\figref{figure:egg-exp}に示す.
  まずCLIPについて, \textbf{ALL}の場合は\secref{subsec:water-exp}や\secref{subsec:butter-exp}と同様に大きく値が波打っており, $E$も小さく$t_{diff}$も大きい.
  \textbf{ONE}や\textbf{OPT}については, $D_{opt}$ではそれなりの性能が確認できており$t_{diff}$も小さい.
  一方で, $D_{eval}$では$D_{opt}$の際と大きく異なる結果を示しており, 非常に精度が低い.
  ImageBindについては, CLIPの場合のように$D_{opt}$と$D_{eval}$で大きな結果の差があるわけではないが, どれも, 状態変化初期に大きく類似度が変化し, その後一定となるケースがほとんどである.
  そのため, どのケースでもうまく状態変化を検知できず, $t_{diff}$は非常に大きい.
  目玉焼きは, 最初に白身の部分が透明から一気に白くなるため, その後の状態変化が非常にわかりにくい料理と言える.
}%

\subsection{Onion Stir-frying Experiment} \label{subsec:onion-exp}
\switchlanguage%
{%
  The results of the onion stir-frying experiment are shown in \figref{figure:onion-exp}.
  Unlike the previous experiments, the number of images is small, so there is no ``average'' and only ``raw'' is shown.
  As for CLIP, the changes in similarity of \textbf{ALL} and \textbf{OPT} are stable as the state changes, indicating that the recognition is highly accurate.
  On the other hand, for \textbf{ONE}, the change in similarity is not as clear as for \textbf{ALL} and \textbf{OPT}, and $t_{diff}$ is larger.
  As for ImageBind, the recognition performance is not so high for \textbf{ALL} and \textbf{ONE}, since the change in similarity fluctuates.
  On the other hand, \textbf{OPT} recognizes the state change with high accuracy as with CLIP.
  Note that the top 5 text prompts and their weights are \textit{cooked onion in this frying pan} (0.36), \textit{raw onion in that frying pan} (0.19), \textit{grilled onion in this frying pan} (0.11), \textit{cooked onion in frying pan} (0.1), and \textit{fresh onion in frying pan} (0.08).
}%
{%
  玉ねぎのソテー実験を行った結果を\figref{figure:onion-exp}に示す.
  これまでの実験とは異なり画像の取得回数が非常に少ないため, ``average''はなく, ``raw''のみを示している.
  まずCLIPについて, \textbf{ALL}と\textbf{OPT}では, 状態変化に応じて類似度が安定して変化しており, 高い精度での認識ができている.
  その一方で\textbf{ONE}では, 類似度変化が\textbf{ALL}や\textbf{OPT}ほどははっきりしておらず, $t_{diff}$が大きくなってしまっている.
  ImageBindについては, \textbf{ALL}と\textbf{ONE}では値が大きく波打っており, 認識性能はあまり高いとは言えない.
  一方で\textbf{OPT}では, CLIPと同様に高い精度で状態変化を認識することができている.
}%

\begin{figure*}[t]
  \centering
  \includegraphics[width=1.9\columnwidth]{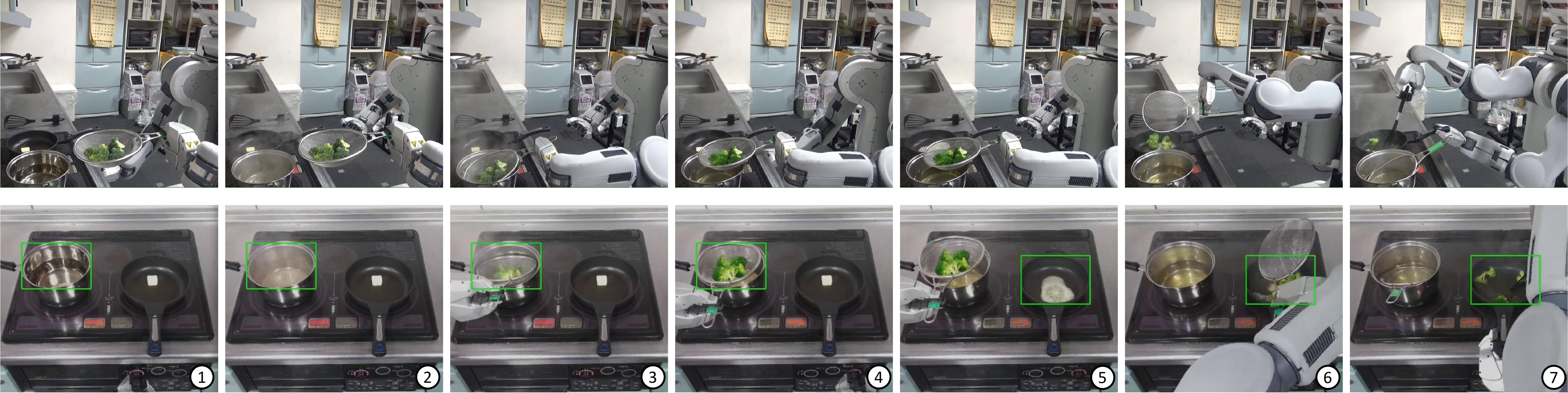}
  \vspace{-1.0ex}
  \caption{Results of the cooking experiment. The PR2 robot boils water, blanches broccoli, and stir-fries it with melted butter.}
  \label{figure:cooking-exp}
  \vspace{-3.0ex}
\end{figure*}

\subsection{Cooking Experiment} \label{subsec:cook-exp}
\switchlanguage%
{%
  The experimental result is shown in \figref{figure:cooking-exp}.
  We place a pot with water and a frying pan with butter on the stove.
  \ctext{1} The robot PR2 turns on the heat for the pot, and \ctext{2} when boiling is detected using the proposed method (ImageBind with \textbf{OPT}), \ctext{3} adds the broccoli in a sieve to the pot.
  After boiling for 3 minutes, \ctext{4} the robot takes the broccoli out, turns on the heat for the frying pan, and \ctext{5} when the proposed method detects that the butter has melted, \ctext{6} adds the broccoli.
  Finally, \ctext{7} the robot stir-fries the boiled broccoli in the frying pan and then turns off the heat.
  A series of cooking behaviors using the proposed method was realized.
}%
{%
  実験結果を\figref{figure:cooking-exp}に示す.
  コンロに水の入った鍋とバターを入れたフライパンを置く.
  ロボットPR2が鍋に火をつけ, 提案手法(ImageBind with\textbf{OPT})によって沸騰を検知したら, ザルに入れたブロッコリーを鍋に入れる.
  次に, 3分茹でたらブロッコリーを取り出す.
  フライパンに火をつけ, 提案手法によってバターが溶けたことを検知したら, ブロッコリーを投入する.
  茹でたブロッコリーをフライパンで炒め, 火を止めた.
  よって, 提案手法を活用した一連の調理行動を実現した.
}%

\section{Discussion} \label{sec:discussion}
\switchlanguage%
{%
  The obtained experimental results are summarized, and their properties and limitations are discussed.
  In the experiments, we handled state changes related to water boiling, butter melting, egg cooking, and onion stir-frying, each of which has different properties, and a variety of results were obtained.
  The state change of water boiling is similar to (ii) in \figref{figure:func-shape}, and is easy to be recognized because the change of state occurs at once towards the end of the process.
  The state change of butter melting is close to (iv) and that of onion stir-frying is close to (i), and both of them can be recognized with high performance for the same reason as above.
  On the other hand, the state change of egg cooking is close to (iii), and the performance is limited because a large state change occurs at the beginning and the subsequent state changes are difficult to be recognized.
  More specifically, the color of the egg white first turns from transparent to white at once, and subsequently there is only a slight color change in egg yolk.
  CLIP and ImageBind, which were used in this study, were not able to overcome this problem, but we expect that the performance will be improved if VLM capable of detecting more precise changes is developed in the future.
  As an overall trend, we found that \textbf{OPT} with black-box optimization has higher recognition performance than \textbf{ALL} and \textbf{ONE}.
  Although the performance of \textbf{ONE} is somewhat higher than that of \textbf{ALL}, it may be reversed depending on the state to be recognized.
  % In addition, as a result of optimization, the text prompts primarily used often employ negative forms, except in the case of onion stir-frying experiments, and there is a tendency for text prompts with different articles in the same state expression to be frequently used.
  Finally, when comparing CLIP and ImageBind, ImageBind shows relatively more stable changes in similarity.
  For CLIP, the results of $D_{opt}$ and $D_{eval}$ are sometimes significantly different from each other, while there are less of these cases for ImageBind.
  On the other hand, there are cases where CLIP performs better than ImageBind, so it is difficult to say which model is better.
  We believe that the simultaneous use of multiple models in the future will improve the performance of continuous state recognition by taking advantage of the characteristics of each model.

  We discuss the prospects of future research.
  First, in this study, continuous state recognition is based only on the correspondence between images and texts, and there are still many unused modalities.
  In particular, video, audio, and heatmaps are indispensable information for cooking, and higher performance can be expected by integrating them into the proposed method.
  Also, in this study, the text set $Q$ is manually created.
  A more practical system can be constructed by obtaining this text set automatically.
  We can use large-scale language models \cite{brown2020gpt3} to obtain multiple synonyms and antonyms of the state to be recognized.
  In addition, we would like to consider various other methods in the future, such as changing the viewing area for robots to focus on, using multiple models described above at the same time, and taking into account incomplete images \cite{yuan2023adaptive}.
}%
{%
  本研究で得られた実験結果をまとめ, その特性と限界を考察する.
  実験では水の沸騰・バターの溶け具合・目玉焼きの固まり具合・玉ねぎのソテー具合に関する状態変化を扱ったが, それぞれ異なる性質を持っており, 多様な結果が得られた.
  水の沸騰は\figref{figure:func-shape}の(ii)に近く, 途中から状態変化が一気に起こるため認識しやすい.
  バターの溶け具合は(iv), 玉ねぎのソテー具合は(i)に近く, これらも高い性能で認識することができた.
  一方で, 目玉焼きの固まり具合は(iii)に近く, 画像的には最初に大きな状態変化が起き, その後の状態変化が非常に分かりにくいため, 状態変化認識には限界があった.
  本研究課で扱ったCLIPとImageBindでは対応することができなかったが, 今後より精密な変化を検知可能な大規模モデルが発展すれば, 性能の向上が期待できる.
  また, 全体的な傾向として, 最適化を行う\textbf{OPT}は\textbf{ALL}や\textbf{ONE}に比べて高い認識性能を有していることが分かった.
  \textbf{ONE}は\textbf{ALL}に比べ多少性能は高いが, 認識する状態によっては逆転することもある.
  % 加えて, 最適化の結果主に用いられるプロンプトは, 玉ねぎのソテー具合実験以外では否定形が多く用いられており, また同じ状態表現で冠詞の異なるプロンプトが多く用いられるという傾向があった.
  最後に, CLIPとImageBindの比較であるが, 比較的類似度変化が安定していたのはImageBindであった.
  CLIPは$D_{opt}$と$D_{eval}$の結果が大きく異なる場合があり, ImageBindはそれが少ない.
  一方で, CLIPの方が認識性能が高い場合もあり, 一概にどちらが優れているとは言うことができない.
  今後複数のモデルを同時に使うことで, それぞれのモデルの特性を活かし, 状態変化認識の性能を向上させることができると考えられる.

  今後の研究課題について述べる.
  まず, 本研究では画像と言語の対応のみから連続状態認識を行っており, 用いることができていないモーダルも多い.
  特に, 動画や音声, ヒートマップ等は調理に欠かせないと情報であり, これらを統合した連続状態認識を行うことで, より高い性能が期待できる.
  また, 本研究では入力するテキスト集合を人手で作成しているが, これを自動的に得ることで, より実用的なシステムを構築することができる.
  大規模言語モデル\cite{brown2020gpt3}により, 認識したい状態の類義語, 対義語を複数回答させることで, 自動的にテキスト集合を獲得しても良い.
  その他, 鍋の中身だけやコンロ全体など見る領域を変化させたり, 上記で述べた複数のモデルを同時に使ったり, 画像の欠損を補ったり\cite{yuan2023adaptive}等, 様々な方法について今後考えていきたい.
}%

\section{CONCLUSION} \label{sec:conclusion}
\switchlanguage%
{%
  In this study, we proposed a continuous state recognition method for cooking robots based on the spoken language through pre-trained large-scale vision-language models.
  A set of texts related to the state to be recognized is prepared, and the similarity between the current image and texts is calculated in the temporal direction.
  In order to make the changes in similarity easier to use for state recognition, we adjusted the weighting of each text based on black-box optimization by fitting it to a sigmoid function and calculating its evaluation value.
  The sigmoid function is suitable for continuous state recognition because it can handle various patterns of state changes.
  The recognition performance with optimization is much better than that without optimization, and we succeeded in recognizing the states of water boiling, butter melting, and onion stir-frying.
  On the other hand, the recognition of egg cooking was difficult due to the fact that a large change in the image occurs in the early stage of the recognition, and the subsequent smaller changes are difficult to be recognized.
  We used CLIP and ImageBind as large-scale vision-language models, and while ImageBind produces more stable recognition results over all, each model has different strengths and weaknesses, and we may consider using both models in combination in the future.
}%
{%
  本研究では, 事前学習済みの大規模視覚-言語モデルにより, 調理ロボットに向けた, 言語に基づく連続的な食材の状態認識を提案した.
  認識したい状態に関するテキスト集合を用意し, それら言語と現在画像間の類似度を計算可能なモデルを用いて, これを時間方向に適用する.
  その類似度変化をより状態認識に用いやすくするため, シグモイド関数へのフィッティングと評価値計算を行い, ブラックボックス最適化に基づいて各テキストの重みを調整した.
  シグモイド関数は様々な状態変化のパターンに対応可能であり, 連続状態認識に適している.
  これにより, 最適化を行わない場合に比べて格段に認識性能が向上し, 水の沸騰・バターの溶け具合・玉ねぎのソテー具合に関する状態認識に成功した.
  一方で, 目玉焼きの固まり具合は画像変化が初期に大きく起こり, その後の変化が分かりにくいため認識が難しいといった限界も分かった.
  また, 大規模言語モデルとしてCLIPとImageBindを用いたが, ImageBindの方が安定した認識結果を得ることができる一方で, 各モデルの得意不得意が異なるため, 今後これらを併用することが考えられる.
}%

{
  %\footnotesize
  %\small
  %\bibliographystyle{junsrt}
  \bibliographystyle{IEEEtran}
  \bibliography{main}
}

\end{document}